\pdfoutput=1

\documentclass[11pt,table,xcdraw]{article}

\usepackage[]{acl}

\usepackage{times}
\usepackage{latexsym}
\usepackage{amsmath}

\usepackage[T1]{fontenc}

\usepackage[utf8]{inputenc}

\usepackage{microtype}

\usepackage{inconsolata}

\usepackage{graphicx}
\usepackage{subcaption}
\usepackage{amsfonts}
\usepackage{booktabs}
\usepackage{multirow}
\usepackage{colortbl}
\usepackage{adjustbox}

\title{Dual Decomposition of Weights and Singular Value Low Rank Adaptation}

\author{
\textbf{Jialong Han\textsuperscript{1,2}},
\textbf{Si Zhang\textsuperscript{2}},
\textbf{Ke Zhang\textsuperscript{2}},
\\
\\
\textsuperscript{1}SIST, ShanghaiTech University\\
\textsuperscript{2}SKLP, Institute of Computing Technology, Chinese Academy of Sciences
}

\begin{document}
\maketitle
\begin{abstract}
        Parameter-Efficient Fine-Tuning (PEFT) has emerged as a critical paradigm for adapting Large Language Models (LLMs) to downstream tasks, 
        among which Low-rank Adaptation (LoRA) represents one of the most widely adopted methodologies. 
        However, existing LoRA-based approaches exhibit two fundamental limitations: 
        unstable training dynamics and inefficient knowledge transfer from pre-trained models, 
        both stemming from random initialization of adapter parameters. 
        To overcome these challenges, we propose DuDe, 
        a novel approach that decomposes weight matrices into magnitude and direction components, 
        employing Singular Value Decomposition (SVD) for principled initialization. 
        Our comprehensive evaluation demonstrates DuDe's superior performance and robustness, 
        achieving up to 48.35\% accuracy on MMLU and 62.53\% (±1.59) accuracy on GSM8K. 
        Our theoretical analysis and empirical validation collectively demonstrate 
        that DuDe's decomposition strategy enhances optimization stability and better preserves pre-trained representations, 
        particularly for domain-specific tasks requiring specialized knowledge. 
        The combination of robust empirical performance and rigorous theoretical foundations establishes DuDe as a significant contribution to PEFT methodologies for LLMs.
\end{abstract}

\section{Introduction}
Pre-trained models have demonstrated exceptional capabilities across diverse applications 
from Natural Language Processing (NLP) tasks \citep{qin-etal-2023-cross} to multi-modal scenarios \citep{li-etal-2023-factual, liu-etal-2023-visual}. 
However, fine-tuning these large models remains computationally expensive.

Parameter-Efficient Fine-Tuning (PEFT) methods have emerged as a promising solution to this challenge.
In particular, Low-Rank Adaptation (LoRA) has gained significant attention due to its ability 
to maintain the model's original architecture while enabling efficient fine-tuning. 
LoRA achieves this by injecting trainable low-rank matrices into the pre-trained weights, 
significantly reducing the number of parameters that need to be updated.

Despite its widespread adoption, LoRA and its variants face two fundamental challenges:
1) Training instability caused by random initialization,
and 2) Inefficient utilization of pre-trained knowledge.
To address these limitations, we propose DuDe (Dual Decomposition of Weights and Singular Value Low Rank Adaptation), 
which employs dual decomposition and Singular Value Decomposition (SVD) based initialization.
Our experimental results validate DuDe's effectiveness through:
(1) More stable training across different random seeds with only ±1.59 standard deviation 
(Section \ref{sec:experiment_details_robustness_seed}),
and (2) Superior performance on knowledge-intensive MMLU tasks achieving up to 48.35\% average accuracy
(Section \ref{sec:experiment_details_robustness_rank}).

DuDe combines two key technical innovations:
magnitude-direction decomposition inspired by DoRA \citep{dora2024} 
and SVD-based initialization building on PiSSA \citep{pissa2024}.
Our main contributions include:

\begin{itemize}
        \item A novel dual decomposition strategy that separates weights into magnitude and direction components, enabling more stable optimization
        \item An SVD-based initialization method that effectively preserves and leverages pre-trained knowledge
        \item Theoretical analysis that demonstrates improved gradient properties and optimization stability
        \item Comprehensive experiments showing consistent performance improvements across diverse models and tasks
\end{itemize}

Our extensive evaluation demonstrates DuDe's strong empirical performance across multiple benchmarks.
Notably, DuDe exhibits exceptional performance on complex tasks requiring domain expertise,
indicating its superior ability to preserve and adapt pre-trained knowledge.

\begin{figure*}[t]
        \begin{subfigure}[b]{0.48\textwidth}
                \centering
                \includegraphics[width=\textwidth]{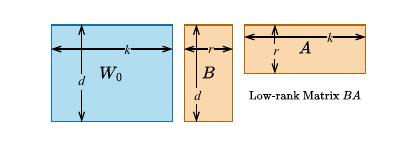}
                \vspace{-1cm}
                \caption{}\label{fig:lora}
        \end{subfigure}
        \begin{subfigure}[b]{0.48\textwidth}
                \centering
                \includegraphics[width=\textwidth]{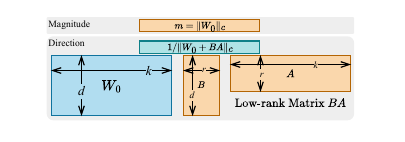}
                \vspace{-1cm}
                \caption{}\label{fig:dora}
        \end{subfigure}
        \caption{The blue parts in the figure represent frozen components, 
        while the orange parts represent trainable components. 
        (a) shows the diagrams of LoRA and PiSSA. 
        The difference between them is that LoRA initializes matrix $B\in\mathbb{R}^{d\times r}$ to 0 
        and matrix $A\in\mathbb{R}^{r\times d}$ to Kaiming uniform distribution, 
        while PiSSA first performs SVD on matrix $W_0$ to obtain $W_0=U\Sigma V^\top$, 
        then sets $B=U_r\sqrt{\Sigma_r}$, $A=\sqrt{\Sigma_r}V_r^\top$, and $W_0=W_0-BA$.
        (b) shows the diagrams of DoRA and DuDe. $m\in\mathbb{R}^{k}$ is the magnitude vector. 
        For the direction matrix, DoRA initializes matrices $B$ and $A$ in the same way as LoRA,
        while DuDe initializes matrices $B$ and $A$ in the same way as PiSSA.
        }
        \label{fig:comparison}
\end{figure*}

\section{Related Work}

Large Language Models (LLMs) containing billions of parameters pose substantial challenges 
in terms of complexity and computational resources when adapting them to new tasks.
PEFT \citep{parameterefficient2019} offers an attractive approach 
by reducing the number of parameters to be fine-tuned and memory requirements, 
while maintaining performance comparable to full fine-tuning.

Existing PEFT methods can be broadly categorized into three main approaches:
Adapter-based Methods \citep{parameterefficient2019,conditional2023,krona2022}, 
Selective Tuning Methods \citep{ben-zaken-etal-2022-bitfit, liao-etal-2023-parameter}, and Re-parameterization Methods.

\paragraph{Re-parameterization Methods} transform the original parameters into a more efficient representation. 
The most prominent example is LoRA \citep{lora2022}, 
which injects trainable adapters into the pre-trained weight through low-rank decomposition. 
Following LoRA, several improvements have been proposed.
DoRA \citep{dora2024} decomposes the pre-trained weight into magnitude and direction components, 
enhancing both learning capacity and training stability.
PiSSA \citep{pissa2024} initializes the adaptor matrices with the principal components of the pre-trained weight, 
freezing the remaining components in a residual matrix.
OFT \citep{parameterefficient2024} exploits orthogonal factorization for model fine-tuning.
LoRA-XS \citep{loraxs2024} and OLoRA \citep{olora2024} further reduce the number of parameters while maintaining performance.
VeRA \citep{vera2024} introduces vector-based random matrix adaptation for more efficient parameterization.
SVFT \citep{svft2024} uses singular vectors for PEFT, sharing some conceptual similarities with our work.

Our work, DuDe, builds upon these advances by combining the strengths of DoRA's magnitude-direction decomposition with PiSSA's SVD-based initialization. 
Unlike previous methods that either focus on decomposition or initialization separately, 
DuDe integrates both aspects to achieve more stable training and better utilization of pre-trained knowledge. 
The key innovation lies in our dual decomposition approach, 
which not only separates magnitude and direction but also performs SVD to initialize the direction matrix, 
leading to more effective adaptation while maintaining parameter efficiency.

\section{Method}

\begin{table*}[t]
        \setlength{\tabcolsep}{1.5pt}
        \centering
        \begin{tabular}{ccccccccccc}
                \toprule
                \textbf{Model} & \textbf{Method} & \textbf{BoolQ} & \textbf{PIQA} & \textbf{SIQA} & \textbf{HellaSwag} & \textbf{WinoGrande} & \textbf{ARC-e} & \textbf{ARC-c} & \textbf{OBQA} & \textbf{Avg.} \\
                \midrule
                \multirow{4}{*}{Qwen1.5-7B} & LoRA & 83.43 & 72.47 & 44.68 & 71.78 & 61.96 & 87.83 & 77.29 & 75.20 & 71.83 \\
                & DoRA & 83.24 & 70.95 & 44.68 & 71.82 & 61.88 & 88.01 & 77.29 & 76.00 & 71.73 \\
                & PiSSA & 84.04 & 74.32 & 44.73 & 71.53 & 61.64 & 87.65 & 78.31 & 74.20 & 72.05 \\
                &\cellcolor{cyan!20} DuDe & \cellcolor{cyan!20} 84.04 & \cellcolor{cyan!20} 75.57 & \cellcolor{cyan!20} 44.98 & \cellcolor{cyan!20} 71.37 & \cellcolor{cyan!20} 62.98 & \cellcolor{cyan!20} 87.65 & \cellcolor{cyan!20} 78.31 & \cellcolor{cyan!20} 75.40 & \cellcolor{cyan!20} \textbf{72.54} \\
                
                \multirow{4}{*}{Qwen2.5-32B} & LoRA & 89.85 & 90.75 & 46.16 & 92.11 & 79.16 & 97.53 & 93.90 & 89.40 & 84.86 \\
                & DoRA & 90.03 & 90.59 & 46.26 & 92.06 & 79.08 & 97.35 & 93.56 & 89.80 & 84.84 \\
                & PiSSA & 89.76 & 89.61 & 46.62 & 91.91 & 77.66 & 97.53 & 91.86 & 88.60 & 84.19 \\
                &\cellcolor{cyan!20} DuDe & \cellcolor{cyan!20} 90.00 & \cellcolor{cyan!20} 90.26 & \cellcolor{cyan!20} 47.54 & \cellcolor{cyan!20} 92.12 & \cellcolor{cyan!20} 77.82 & \cellcolor{cyan!20} 97.18 & \cellcolor{cyan!20} 93.22 & \cellcolor{cyan!20} 91.00 & \cellcolor{cyan!20} \textbf{84.89} \\
                
                \multirow{4}{*}{LLaMA2-13B} & LoRA & 65.84 & 73.78 & 53.68 & 48.63 & 51.78 & 79.01 & 59.32 & 60.00 & 61.51 \\
                & DoRA & 61.07 & 73.94 & 54.25 & 49.98 & 51.46 & 79.19 & 61.02 & 60.40 & 61.41 \\
                & PiSSA & 66.09 & 70.18 & 45.39 & 51.94 & 52.33 & 82.19 & 59.32 & 62.40 & 61.23 \\
                &\cellcolor{cyan!20} DuDe & \cellcolor{cyan!20} 72.72 & \cellcolor{cyan!20} 74.10 & \cellcolor{cyan!20} 45.75 & \cellcolor{cyan!20} 60.39 & \cellcolor{cyan!20} 51.22 & \cellcolor{cyan!20} 82.54 & \cellcolor{cyan!20} 61.02 & \cellcolor{cyan!20} 62.20 & \cellcolor{cyan!20} \textbf{63.74} \\
                \bottomrule
        \end{tabular}
        \caption{Accuracy comparison of Qwen1.5-7B, Qwen2.5-32B, and LLaMA2-13B with different PEFT methods on eight commonsense reasoning tasks. The best results are highlighted in bold.}
        \label{tab:exp_commonsense_results}
\end{table*}

\subsection{Preliminaries}

Building upon the hypothesis that fine-tuning updates exhibit a low "intrinsic rank" \citep{intrinsic2021}, 
LoRA \citep{lora2022} employs the product of two low-rank matrices to efficiently update pre-trained weights (Figure \ref{fig:lora}). 
For a pre-trained weight matrix $W_0\in\mathbb{R}^{d\times k}$, 
LoRA parameterizes the weight update $\Delta W\in\mathbb{R}^{d\times k}$ 
as a low-rank decomposition $BA$, 
where $B\in\mathbb{R}^{d\times r}$ and $A\in\mathbb{R}^{r\times k}$ 
are low-rank matrices with rank $r\ll\min(d,k)$. 
The fine-tuned weight $W'$ is therefore formulated as: 
\begin{equation}
        W'=W_0+\Delta W=W_0+\underline{BA}
\end{equation}

During fine-tuning, $W_0$ remains frozen while only the low-rank matrices are trained.
The matrices $A$ are initialized using the Kaiming uniform distribution \citep{delving2015}, 
while matrices $B$ are initialized to zero,
ensuring that $\Delta W=BA$ starts from zero at the beginning of training; 
thus the injection of adapters does not affect the model's output initially.

Inspired by \citet{weight2016}, DoRA \citep{dora2024} decomposes the pre-trained weight into magnitude and direction components, 
and fine-tunes both components simultaneously (Figure \ref{fig:dora}). 
To efficiently update the directional component with its large parameter space, 
DoRA adopts LoRA's low-rank decomposition approach.
The formulation is expressed as:
\begin{equation}
        W'=\underline{m}\frac{W_0+\Delta W}{\|W_0 + \Delta W\|_c}=\underline{m}\frac{W_0+\underline{BA}}{\|W_0+\underline{BA}\|_c}
\end{equation}
where $m\in\mathbb{R}^{k}$ represents the trainable magnitude vector, 
$\Delta W=BA$ is the directional update parameterized by two low-rank matrices $B\in\mathbb{R}^{d\times r}$ and $A\in\mathbb{R}^{r\times k}$ (with $r\ll\min(d,k)$), 
$\|\cdot\|_c$ denotes the column-wise vector norm,
and underlined parameters are trainable during fine-tuning.
Following LoRA, matrices $B$ and $A$ are initialized to ensure $\Delta W=0$ at the beginning of training, 
maintaining the model's initial behavior while enabling effective adaptation.

\subsection{Dual Decomposition of Weights and Singular Value Low Rank Adaptation}
In this section, we present our proposed method, 
Dual Decomposition of Weights and Singular Value Low Rank Adaptation (DuDe).
As illustrated in Figure \ref{fig:comparison}, 
DuDe performs SVD on the pre-trained weight matrix $W_0$ to derive optimal initialization parameters for low-rank adaptation.
When applying SVD to a matrix $W_0\in\mathbb{R}^{d\times k}$, 
we obtain the decomposition $W_0=U\Sigma V^\top$, 
where $U\in\mathbb{R}^{d\times p}$ and $V\in\mathbb{R}^{k\times p}$ are orthogonal matrices containing the left and right singular vectors, 
and $\Sigma\in\mathbb{R}^{p\times p}$ is a diagonal matrix containing the singular values of $W_0$ in descending order,
with $p = \min(d,k)$.

To effectively capture the most important features, 
the top $r$ singular values and their corresponding singular vectors are extracted from $\Sigma$, $U$, and $V$,
which are denoted as $\Sigma_r\in\mathbb{R}^{r\times r}$, $U_r\in\mathbb{R}^{d\times r}$, and $V_r\in\mathbb{R}^{k\times r}$.
These components form the update matrix:
\begin{equation}
        \Delta W=U_r\Sigma_rV_r^\top
\end{equation}

The remaining components of the original weight matrix are preserved as:
\begin{equation}
        W_f = W_0 - \Delta W
\end{equation}
where $W_f$ remains frozen during fine-tuning.

The low-rank matrices are initialized using the SVD components for efficient parameterization:
\begin{equation}
        A = \sqrt{\Sigma_r}V_r^\top \in \mathbb{R}^{r\times k}
\end{equation}
\begin{equation}
        B = U_r\sqrt{\Sigma_r} \in \mathbb{R}^{d\times r}
\end{equation}
where $B$ and $A$ are low-rank matrices with rank $r\ll p$.

The final fine-tuned weight $W'$ integrates the frozen component $W_f$ with the trainable low-rank update, scaled by a trainable magnitude vector $m$:
\begin{equation}
        W' = \underline{m}\frac{W_f + \Delta W}{\|W_f + \Delta W\|_c}
        = \underline{m}\frac{W_f + \underline{BA}}{\|W_f + \underline{BA}\|_c}
        \label{eq:dude}
\end{equation}
At initialization, since $\Delta W=BA$, the fine-tuned weight $W'$ is equivalent to the original weight $W_0$, ensuring that the model's initial behavior is preserved while enabling effective adaptation during training.

\subsection{Gradient Analysis}
In this section, we analyze the gradient of DuDe and demonstrate how our proposed decomposition enables more stable and efficient fine-tuning.

From Eq. \eqref{eq:dude}, the gradient of loss $\mathcal{L}$ with respect to $m$ and $W_0=W_f+\Delta W$ can be derived as:
\begin{equation}
        \frac{\partial \mathcal{L}}{\partial m} = 
        \frac{\partial \mathcal{L}}{\partial W'}\frac{W_0}{\|W_0\|_c}
        \label{eq:grad_m}
\end{equation}
\begin{equation}
        \frac{\partial \mathcal{L}}{\partial W_0} = 
        \frac{m}{\|W_0\|_c}\left(
        \mathbb{I}-\frac{W_0 W_0^\top}{\|W_0\|_c^2}
        \right)
        \frac{\partial \mathcal{L}}{\partial W'}
        \label{eq:grad_w0}
\end{equation}

Eq. \eqref{eq:grad_w0} reveals that the gradient of $W_0$ undergoes two key transformations: scaling by $\frac{m}{\|W_0\|_c}$ and projection onto the orthogonal complement of $W_0$. These transformations help align the gradient's covariance matrix more closely with the identity matrix $\mathbb{I}$, promoting optimization stability.

Since $W_0=W_f+\Delta W$, the gradient $\frac{\partial \mathcal{L}}{\partial W_0}$ is equivalent to $\frac{\partial \mathcal{L}}{\partial \Delta W}$. Consequently, all optimization benefits from this decomposition directly transfer to $\Delta W$, enhancing DuDe's learning stability.

Furthermore, because the top $r$ singular values and their corresponding singular vectors capture the most significant features of $W_0$, the gradient $\frac{\partial \mathcal{L}}{\partial \Delta W}$ contains more stable and informative signals compared to LoRA's gradient, leading to improved convergence properties.

\begin{figure*}[t]
        \centering
        \begin{subfigure}[b]{0.45\textwidth}
                \centering
                \includegraphics[width=\textwidth]{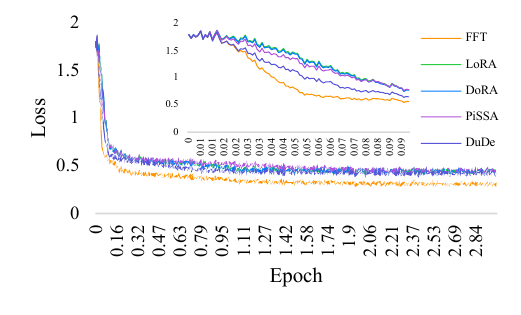}
                \caption{Training loss}
                \label{fig:exp_different_epochs_loss}
        \end{subfigure}
        \begin{subfigure}[b]{0.45\textwidth}
                \centering
                \includegraphics[width=\textwidth]{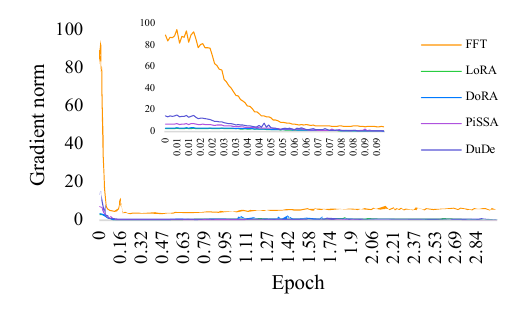}
                \caption{Gradient norm}
                \label{fig:exp_different_epochs_grad_norm}
        \end{subfigure}
        \begin{subfigure}[b]{0.9\textwidth}
                \centering
                \includegraphics[width=\textwidth]{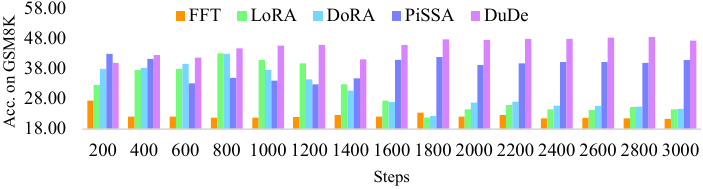}
                \caption{Accuracy}
                \label{fig:exp_different_epochs_acc}
        \end{subfigure}
        \caption{Comparison of Full finetuning, DuDe and other PEFT methods on Mistral 7B v0.2 model: (a) Training loss, (b) Gradient norm during training on MetaMathQA-395K dataset for 3 epochs, and (c) Evaluation accuracy on GSM8K dataset measured every 200 steps over 3000 total training steps.}
        \label{fig:exp_different_epochs}
\end{figure*}

Our experiments, as illustrated in Figure \ref{fig:exp_different_epochs_loss}, show that DuDe's loss and gradient norm curves closely resemble those of full fine-tuning, confirming that our dual decomposition effectively transfers the benefits of full fine-tuning while maintaining parameter efficiency.

\begin{table}[t]
        \centering
        \begin{tabular}{ccc}
                \toprule
                \textbf{Model} & \textbf{Method} & \textbf{Score} \\
                \midrule
                \multirow{4}{*}{Qwen1.5-7B} & LoRA & 20.20 \\
                & DoRA & 22.22 \\
                & PiSSA & 19.19 \\
                & \cellcolor{cyan!20} DuDe & \cellcolor{cyan!20} \textbf{24.75} \\
                \multirow{4}{*}{Qwen2.5-14B} & LoRA & 39.39 \\
                & DoRA & 40.40 \\
                & PiSSA & 40.91 \\
                & \cellcolor{cyan!20} DuDe & \cellcolor{cyan!20} \textbf{41.41} \\
                \multirow{4}{*}{Mistral-7B v0.1} & LoRA & 15.66 \\
                & DoRA & 20.20 \\
                & PiSSA & 20.71 \\
                & \cellcolor{cyan!20} DuDe & \cellcolor{cyan!20} \textbf{23.74} \\
                \multirow{4}{*}{Phi4 small} & LoRA & 30.81 \\
                & DoRA & 33.33 \\
                & PiSSA & 35.35 \\
                & \cellcolor{cyan!20} DuDe & \cellcolor{cyan!20} \textbf{39.90} \\
                \bottomrule
        \end{tabular}
        \caption{Score comparison of Qwen1.5-7B, Qwen2.5-14B, Mistral-7B v0.1, and Phi4 small with different PEFT methods on GPQA task. The best results are highlighted in bold.}
        \label{tab:exp_gpqa_results}
\end{table}

\section{Experiments}

\subsection{Commonsense Reasoning}
DuDe is comprehensively evaluated against established PEFT methods (LoRA, DoRA, and PiSSA) 
on commonsense reasoning tasks across three different models: 
Qwen1.5-7B \citep{qwen1.5}, Qwen2.5-32B \citep{qwen252025}, and LLaMA2-13B \citep{llama2023}. 
The evaluation suite is comprised of eight diverse commonsense reasoning benchmarks:
BoolQ \citep{boolq2019}, 
PIQA \citep{piqa2019}, 
SIQA \citep{sap-etal-2019-social},
HellaSwag \citep{hellaswag2019}, 
Winogrande \citep{winogrande2021}, 
ARC-e/ARC-c \citep{think2018}, 
and OpenBookQA \citep{can2018}.
For all experiments, CommonsenseQA \citep{talmor-etal-2019-commonsenseqa} is used for fine-tuning and evaluations are performed on the respective test sets using the OpenCompass \citep{opencompass2023} framework.
For fair comparison, identical hyperparameters including rank $r$, learning rate, batch size, and training epochs are shared across all methods, with details being provided in Table \ref{tab:exp_commonsense_config}.

As shown in Table \ref{tab:exp_commonsense_results}, 
DuDe consistently outperforms all baseline methods across all three models. 
For Qwen1.5-7B, DuDe achieves an average accuracy of 72.54\%, surpassing LoRA (71.83\%), DoRA (71.73\%), 
and PiSSA (72.05\%), with particularly strong improvements on PIQA (+3.10\% over LoRA) and Winogrande (+1.02\% over LoRA). 
On Qwen2.5-32B, DuDe maintains its advantage with 84.89\% average accuracy, 
showing notable gains on SIQA (+1.38\% over LoRA). 
The most substantial improvements appear with LLaMA2-13B, where DuDe achieves 63.74\% average accuracy, 
significantly outperforming LoRA (61.51\%) by 2.23\%. 
In this case, DuDe demonstrates remarkable gains on HellaSwag (+11.76\% over LoRA) and BoolQ (+6.88\% over LoRA), 
highlighting its effectiveness in adapting different models to commonsense reasoning tasks.

\subsection{GPQA Task}\label{sec:experiment_details_gpqa}
In this section, DuDe is evaluated on the GPQA \citep{gpqa2024} dataset,
a challenging benchmark of graduate-level questions in biology, physics, and chemistry that cannot be easily answered through online searches. Deep domain knowledge and sophisticated reasoning capabilities are required by these questions.

Four different models (Qwen1.5-7B, Qwen2.5-14B, Mistral-7B v0.1 \citep{mistral2023}, and Phi4 small \citep{phi42024}) are fine-tuned on both the \textsf{Main} and \textsf{Extended} splits of GPQA,
and their performance is evaluated on the \textsf{Diamond} split using the OpenCompass framework.
Similar to the commonsense reasoning experiments,
identical hyperparameters are maintained across all PEFT methods (LoRA, DoRA, PiSSA, and DuDe),
including rank $r$, learning rate, and training epochs. However, due to the complexity of GPQA,
a smaller batch size (set to 4) is used compared to the commonsense tasks.
For Phi4 small model, $W_{qkv}$ is used as the target module due to its different architecture,
while the same target modules as in commonsense experiments are maintained for the other models.

Table \ref{tab:exp_gpqa_results} presents our findings. 
DuDe consistently outperforms all baseline methods across all models tested. 
With Qwen1.5-7B, DuDe achieves 24.75\%, significantly surpassing LoRA (20.20\%), DoRA (22.22\%), and PiSSA (19.19\%). 
On Qwen2.5-14B, DuDe reaches 41.41\%, maintaining a consistent advantage over the baselines. 
For Mistral-7B v0.1, DuDe scores 23.74\%, outperforming LoRA by a substantial 8.08 percentage points. 
The most dramatic improvement appears with Phi4 small, 
where DuDe achieves 39.90\%, exceeding LoRA (30.81\%) by 9.09 percentage points.

These results demonstrate DuDe's effectiveness in adapting various model architectures to complex, 
knowledge-intensive tasks requiring specialized expertise. 
The consistent performance improvements across different models highlight DuDe's versatility 
and robustness as a PEFT method.

\begin{table*}[t]
        \centering
        \setlength{\tabcolsep}{5pt}
        \begin{tabular}{cccccccc}
                \toprule
                \textbf{rank $r$} & \textbf{PEFT Method} & \textbf{Humanities} & \textbf{Social Science} & \textbf{STEM} & \textbf{Other} & \textbf{Avg.} & \textbf{Weighted Avg.} \\
                \midrule
                \multirow{4}{*}{2} & LoRA & 49.09 & 51.87 & 40.54 & 47.74 & 46.51 & 45.11\\
                & DoRA & 48.88 & 51.95 & 40.17 & 47.39 & 46.28 & 44.83 \\
                & PiSSA & 49.16 & 52.24 & 41.86 & 48.60 & \textbf{47.24} & 45.53 \\
                & \cellcolor{cyan!20} DuDe & \cellcolor{cyan!20} 49.06 & \cellcolor{cyan!20} 52.48 & \cellcolor{cyan!20} 41.57 & \cellcolor{cyan!20} 48.37 & \cellcolor{cyan!20} 47.13 & \cellcolor{cyan!20} \textbf{45.56} \\
                \midrule
                \multirow{4}{*}{4} & LoRA & 49.28 & 51.38 & 40.05 & 47.11 & 46.15 & 44.93 \\
                & DoRA & 48.87 & 52.62 & 40.19 & 47.62 & 46.48 & 45.05 \\
                & PiSSA & 49.28 & 51.77 & 40.62 & 47.24 & 46.45 & 44.98 \\
                & \cellcolor{cyan!20} DuDe & \cellcolor{cyan!20} 49.94 & \cellcolor{cyan!20} 52.34 & \cellcolor{cyan!20} 40.57 & \cellcolor{cyan!20} 47.46 & \cellcolor{cyan!20} \textbf{46.75} & \cellcolor{cyan!20} \textbf{45.34} \\
                \midrule
                \multirow{4}{*}{8} & LoRA & 47.91 & 51.02 & 38.30 & 47.66 & 45.30 & 43.68 \\
                & DoRA & 47.65 & 50.79 & 36.47 & 46.87 & 44.40 & 43.09 \\
                & PiSSA & 48.65 & 52.02 & 39.22 & 47.24 & 45.90 & 44.21 \\
                & \cellcolor{cyan!20} DuDe & \cellcolor{cyan!20} 49.07 & \cellcolor{cyan!20} 52.08 & \cellcolor{cyan!20} 40.01 & \cellcolor{cyan!20} 47.12 & \cellcolor{cyan!20} \textbf{46.24} & \cellcolor{cyan!20} \textbf{44.72} \\
                \midrule
                \multirow{4}{*}{16} & LoRA & 48.64 & 51.17 & 41.48 & 47.32 & 46.48 & 44.99\\
                & DoRA & 49.44 & 53.12 & 39.86 & 48.37 & 46.78 & 45.41 \\
                & PiSSA & 50.00 & 52.78 & 41.35 & 47.50 & 47.13 & 45.57 \\
                & \cellcolor{cyan!20} DuDe & \cellcolor{cyan!20} 50.11 & \cellcolor{cyan!20} 53.04 & \cellcolor{cyan!20} 41.50 & \cellcolor{cyan!20} 47.84 & \cellcolor{cyan!20} \textbf{47.34} & \cellcolor{cyan!20} \textbf{45.88} \\
                \midrule
                \multirow{4}{*}{32} & LoRA & 49.71 & 52.63 & 39.97 & 48.52 & 46.81 & 45.48 \\
                & DoRA & 49.78 & 52.82 & 40.49 & 47.83 & 46.88 & 45.48\\
                & PiSSA & 50.12 & 51.90 & 40.96 & 47.86 & 46.92 & 45.45 \\
                & \cellcolor{cyan!20} DuDe & \cellcolor{cyan!20} 50.44 & \cellcolor{cyan!20} 53.84 & \cellcolor{cyan!20} 42.85 & \cellcolor{cyan!20} 49.26 & \cellcolor{cyan!20} \textbf{48.35} & \cellcolor{cyan!20} \textbf{46.52}\\
                \bottomrule
        \end{tabular}
        \caption{Comparison of the average accuracy between LoRA and DuDe method across various rank settings for MMLU tasks. 
        DuDe consistently outperforms LoRA at all rank settings. We also compare DuDe with DoRA and PiSSA, 
        and find that DuDe achieves better performance than DoRA and PiSSA at all rank settings.
        The best results are highlighted in bold.}
        \label{tab:exp_different_rank_settings}
\end{table*}

\subsection{Robustness to Different Epochs Settings}
In this section, Mistral-7B v0.2 model is finetuned on MetaMathQA-395K \citep{metamath2024} dataset.
The detailed configuration is shown in Table \ref{tab:exp_robustness_epochs_config}.
The training loss and gradient norms are visualized and evaluated on the GSM8K \citep{cobbe2021training} dataset every 200 steps, 
by which quicker convergence and superior performance of DuDe compared to other PEFT methods are demonstrated.

As shown in Figure \ref{fig:exp_different_epochs}, DuDe demonstrates superior performance compared to other PEFT methods across multiple metrics. From the training loss curve in Figure \ref{fig:exp_different_epochs_loss}, we observe that DuDe converges more quickly compared to LoRA, DoRA, and PiSSA. This faster convergence can be attributed to DuDe's dual decomposition approach and SVD-based initialization, which provides a better starting point for optimization.

Most notably, the accuracy plot in Figure \ref{fig:exp_different_epochs_acc} demonstrates DuDe's consistent performance advantage. Starting from early training steps, DuDe achieves higher accuracy on the GSM8K evaluation set and maintains this lead throughout the training process. By the end of training, DuDe reaches a significantly higher final accuracy compared to baseline methods, indicating better generalization capabilities.

These empirical results validate our theoretical analysis that DuDe's decomposition strategy leads to more stable optimization dynamics and better utilization of the pre-trained model's knowledge. The combination of magnitude-direction decomposition and SVD-based initialization appears to create a more favorable optimization landscape, resulting in both faster convergence and superior final performance.

\begin{figure}[t]
        \centering
        \begin{subfigure}[b]{0.45\textwidth}
                \centering
                \includegraphics[width=\textwidth]{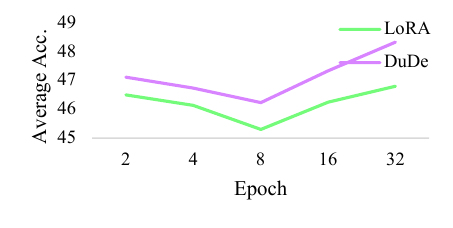}
                \caption{Accuracy}
        \end{subfigure}
        \begin{subfigure}[b]{0.45\textwidth}
                \centering
                \includegraphics[width=\textwidth]{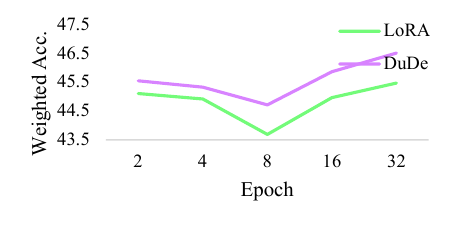}
                \caption{Weighted Average Accuracy}
        \end{subfigure}
        \caption{Performance comparison between LoRA and DuDe on MMLU tasks with varying rank settings. 
        (a) Average accuracy across all MMLU categories shows DuDe consistently outperforming LoRA, especially at larger ranks. 
        (b) Weighted average accuracy demonstrates similar trends, with DuDe maintaining superior performance across all rank configurations.}
        \label{fig:exp_different_rank_settings}
\end{figure}

\begin{figure}[t]
        \centering
        \includegraphics[width=0.48\textwidth]{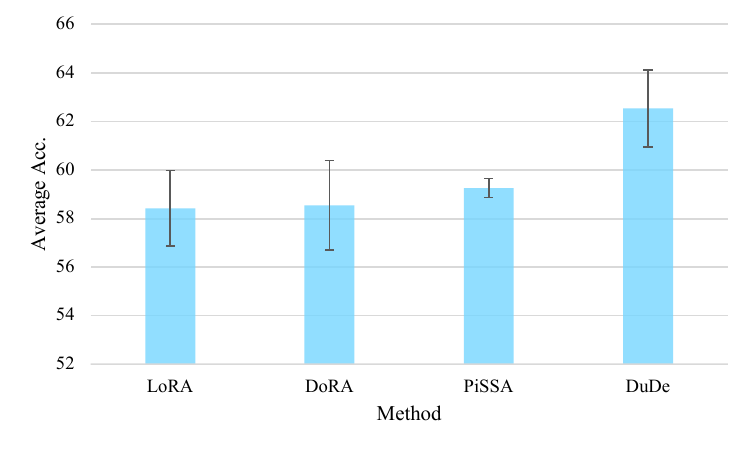}
        \caption{Average accuracy of DuDe and LoRA on MMLU tasks with different seeds.}
        \label{fig:exp_different_seeds}
\end{figure}

\subsection{Robustness to Different Rank Settings}\label{sec:experiment_details_robustness_rank}
In this section, how different rank settings affect model performance is investigated by comparing DuDe with other PEFT methods. Experiments are conducted on Qwen2.5-0.5B using MMLU tasks \citep{measuring2021a}, 
where the rank $r$ is varied among $\{2, 4, 8, 16, 32\}$. 
The detailed configuration is presented in Table \ref{tab:exp_robustness_rank_config}.
The results are presented in Figure \ref{fig:exp_different_rank_settings} and Table \ref{tab:exp_different_rank_settings}.

As illustrated in Figure \ref{fig:exp_different_rank_settings}, DuDe demonstrates consistently superior performance across all rank configurations. The performance advantage becomes more pronounced as rank increases, with DuDe achieving the best results at $r=32$ (48.35\% average accuracy and 46.52\% weighted average accuracy). This represents improvements of 1.54\% and 1.04\% over LoRA respectively.

A detailed analysis of Table \ref{tab:exp_different_rank_settings} reveals several key findings: 1) Performance Scaling: DuDe shows better scaling with increased rank compared to baseline methods. At $r=32$, DuDe achieves the highest scores across all categories, with particularly strong performance in STEM (42.85\%) and humanities (50.44\%) subjects. 2) Low-Rank Efficiency: At lower ranks ($r=2,4$), while all methods perform similarly due to limited parameter capacity, DuDe maintains a slight advantage in weighted average accuracy (45.56\% at $r=2$, 45.34\% at $r=4$).

These results show that DuDe's dual decomposition and initialization strategies enable better model capacity utilization and achieve more robust performance across different ranks.

\subsection{Robustness to Different Seed Settings}\label{sec:experiment_details_robustness_seed}
In this section, a comprehensive analysis of DuDe's robustness across different random seed settings is conducted. Qwen1.5-7B is finetuned on GSM8K tasks using five different random seeds (42, 78, 512, 1234, 3407).

The detailed performance trajectory across different seeds is visualized in Figure \ref{fig:exp_different_seeds}, 
which clearly illustrates DuDe's robust advantage over baseline methods.
The experimental results demonstrate DuDe's superior stability and performance. 
Across all five seed settings, DuDe achieves the highest average accuracy of 62.53\% with a standard deviation of 1.59. 
This represents a significant improvement over existing methods:
\paragraph{LoRA} 58.42\% average accuracy (±1.55 std)
\paragraph{DoRA} 58.55\% average accuracy (±1.85 std) 
\paragraph{PiSSA} 59.26\% average accuracy (±0.38 std)

Most remarkably, even DuDe's worst performance (61.11\% with seed 3407) surpasses the best results achieved by all baseline methods (LoRA's best: 60.35\% with seed 42, DoRA's best: 60.96\% with seed 42, PiSSA's best: 59.67\% with seed 521). This demonstrates that DuDe not only achieves higher average performance but also maintains consistently superior results regardless of random initialization. 

\begin{table*}[t]
        \centering
        \begin{tabular}{ccccccccc}
                \toprule
                \textbf{Model} & \textbf{Dataset} & \textbf{Metric} & \textbf{LoRA} & \textbf{DoRA} & \textbf{PiSSA} & \textbf{DuDe} & \textbf{DuDe$_\mathcal{A}$} & \textbf{DuDe$_\mathcal{B}$} \\
                \midrule
                Qwen1.5-7B & GSM8K & Acc. & 60.35 & 60.96 & 59.44 & 64.22 & 67.48 & 66.72 \\
                \multirow{6}{*}{Qwen2.5-0.5B} & \multirow{6}{*}{MMLU} & Humanities & 49.71 & 49.78 & 50.12 & 50.44 & 50.54 & 50.31 \\
                & & Social Science & 52.63 & 52.82 & 51.90 & 53.84 & 53.22 & 53.28 \\
                & & STEM & 39.97 & 40.49 & 40.96 & 42.85 & 42.40 & 42.27 \\
                & & Other & 48.52 & 47.83 & 47.86 & 49.26 & 49.57 & 48.61 \\
                & & Avg. & 46.81 & 46.88 & 46.92 & 48.35 & 48.17 & 47.87 \\
                & & Weighted Avg. & 45.48 & 45.48 & 45.45 & 46.52 & 46.62 & 46.10\\
                \bottomrule
        \end{tabular}
        \caption{Performance comparison of different PEFT methods and DuDe variants on GSM8K and MMLU benchmarks. DuDe$_\mathcal{A}$ and DuDe$_\mathcal{B}$ represent different initialization strategies for the dual decomposition matrices.}
        \label{tab:exp_different_initialization}
\end{table*}

\subsection{Differentiable Initialization}
In this section, how different initialization strategies affect DuDe's performance is investigated.
Specifically, two initialization variants are explored:
\begin{equation}
        A = \Sigma_rV_r^\top, \quad B = U_r
\end{equation}
and 
\begin{equation}
        A = V_r^\top, \quad B = U_r\Sigma_r
\end{equation}
which we denote as DuDe$_\mathcal{A}$ and DuDe$_\mathcal{B}$ respectively. These variants differ in how they distribute the singular values between matrices A and B.

Qwen1.5-7B is finetuned on the MetaMathQA-395K dataset and Qwen2.5-0.5B is finetuned on MMLU tasks using different initialization methods. The experimental results are reported in Table \ref{tab:exp_different_initialization}.

The experimental results reveal several interesting patterns. For the GSM8K dataset using Qwen1.5-7B, both DuDe$_\mathcal{A}$ and DuDe$_\mathcal{B}$ significantly outperform the baseline DuDe implementation, achieving accuracies of 67.48\% and 66.72\% respectively, compared to DuDe's 64.22\%. This suggests that carefully distributing singular values between matrices A and B during initialization can lead to better optimization dynamics.

For the MMLU benchmark using Qwen2.5-0.5B, the performance differences between initialization variants are more nuanced. DuDe$_\mathcal{A}$ shows slight improvements in Humanities (50.54\% vs 50.44\%) and Other categories (49.57\% vs 49.26\%), while performing marginally lower in Social Science (53.22\% vs 53.84\%) and STEM (42.40\% vs 42.85\%) compared to standard DuDe. DuDe$_\mathcal{B}$ generally performs slightly below both DuDe and DuDe$_\mathcal{A}$ across most categories, though the differences are relatively small.

Overall, while both initialization variants demonstrate competitive performance, DuDe$_\mathcal{A}$ appears to be the most promising, achieving the highest weighted average accuracy (46.62\%) on MMLU and the best performance (67.48\%) on GSM8K. This suggests that allocating singular values to matrix A during initialization may provide better optimization properties for PEFT.

\section{Conclusion}
In this paper, we introduced DuDe, 
a novel PEFT approach that combines dual decomposition of weights with singular value low-rank adaptation. 
Our method addresses two key limitations of existing PEFT approaches: training instability and under-utilization of pre-trained knowledge. 
Through the decomposition of weight matrices into magnitude and direction components, 
along with SVD-based initialization, DuDe achieves more stable optimization while better preserving the knowledge encoded in pre-trained models.

Our extensive experimental evaluation demonstrates DuDe's superior performance across multiple dimensions:

\begin{itemize}
        \item Consistent improvements over baseline methods across different rank settings on the MMLU benchmark, achieving up to 48.35\% average accuracy
        \item Robust performance across different random seeds on the GSM8K dataset, with an average accuracy of 62.53\% (±1.59)
        \item Strong performance on complex tasks requiring deep domain expertise, suggesting better preservation of pre-trained knowledge
\end{itemize}

The theoretical analysis and empirical results validate our key design choices, showing how the dual decomposition strategy leads to more stable gradients and better optimization properties. These findings suggest that DuDe represents a meaningful step forward in PEFT, offering a more principled approach to adapting LLMs.

Future work could explore extending DuDe to other model architectures, investigating its effectiveness in multi-task scenarios, and further analyzing the theoretical foundations of its improved stability. Additionally, combining DuDe with other PEFT innovations could potentially yield even more efficient and effective adaptation methods.

\section{Limitations}
Despite DuDe's promising results, several key limitations need to be acknowledged. 
The SVD-based initialization, while effective, 
introduces additional computational overhead during setup compared to simpler methods. 
This one-time cost can become significant when working with extremely large models or when rapid deployment is needed. 
Memory usage is also slightly higher than basic LoRA due to storing both magnitude and direction components, 
which may be problematic in resource-constrained environments.

Our current implementation focuses mainly on transformer architectures, particularly attention layers. 
The method's effectiveness on other architectures or different transformer components, 
remains to be thoroughly explored. 
The optimal application to emerging architectures such as mixture-of-experts models is also unclear.

While DuDe excels at complex tasks requiring domain expertise, 
its advantages may be less pronounced for simpler tasks where standard PEFT methods already perform well. 
This task-dependent variation makes it challenging to provide universal recommendations for its use. 
Additionally, while we offer some theoretical analysis, a complete understanding of why certain initialization strategies outperform others remains incomplete. 
The interaction between magnitude-direction decomposition and SVD-based initialization warrants deeper theoretical investigation.

Our experiments, though comprehensive, primarily focus on models up to 32B parameters. 
Further research is needed to understand DuDe's scaling behavior on larger models (70B+ parameters) 
and its interaction with other scaling laws. 
Future work should focus on developing more efficient initialization methods, 
extending architecture support, deepening theoretical understanding, 
and studying scaling properties in extremely large models.


\bibliography{anthology,custom,ref}

\appendix
\section{Experiment Details}
\label{sec:experiment_details}

\subsection{Commonsense Reasoning}
\label{sec:experiment_details_commonsense}
In this section, we provide the details of the commonsense reasoning experiments. 
We fine-tune the models for one epoch using a batch size of 1 and gradient accumulation steps of 20. 
The learning rate is set to 2e-5 with cosine decay scheduling and 0.03\% warmup rate. 
We apply our method to query and value matrices ($W_q$, $W_v$) in the attention layers with rank $r=16$. 
The detailed hyperparameter settings are shown in Table \ref{tab:exp_commonsense_config}.

\begin{table*}[h]
        \centering
        \begin{tabular}{ccccccc}
                \toprule
                \textbf{rank $r$} & \textbf{learning rate} & \textbf{epochs} & \textbf{warmup \%} & \textbf{scheduler} & \textbf{packing} & \textbf{target module} \\
                \midrule
                16 & 2e-5 & 1 & 0.03 & cosine & false & $W_q, W_v$ \\
                \bottomrule
        \end{tabular}
        \caption{Configuration for commonsense reasoning experiments. Note that the batch size is set to 1 and the gradient accumulation steps is set to 20.}
        \label{tab:exp_commonsense_config}
\end{table*}

\subsection{Settings for Robustness Experiments to Different Epochs}
\label{sec:experiment_details_robustness_epochs}
In this section, we provide the details of the robustness experiments conducted across different training epochs.
We randomly sampled 128,000 examples from the MetaMathQA-395K dataset using a fixed random seed of 42 to ensure reproducibility.
For both PEFT methods and full fine-tuning experiments, we used identical learning rate settings to enable fair comparisons.
Specifically, we trained each model configuration for 3 epochs to analyze the impact of training duration on model performance.
The learning rate was set to 2e-5 with cosine decay scheduling and 0.03\% warmup rate, consistent across all experimental conditions.
The detailed configuration is shown in Table \ref{tab:exp_robustness_epochs_config}.

\begin{table*}[h]
        \centering
        \begin{tabular}{ccccccc}
                \toprule
                \textbf{rank $r$} & \textbf{learning rate} & \textbf{epochs} & \textbf{warmup \%} & \textbf{scheduler} & \textbf{packing} & \textbf{target module} \\
                \midrule
                16 & 2e-5 & 3 & 0.03 & cosine & false & $W_q, W_v$ \\
                \bottomrule
        \end{tabular}
        \caption{Configuration for robustness experiments to different epochs. Note that the batch size is set to 8 and the gradient accumulation steps is set to 16.}
        \label{tab:exp_robustness_epochs_config}
\end{table*}

\subsection{Settings for Robustness Experiments to Different Rank Settings}
\label{sec:experiment_details_robustness_rank_settings}
In this section, we provide the details of the robustness experiments to different rank settings. 
The detailed configuration is shown in Table \ref{tab:exp_robustness_rank_config}.

\begin{table*}[h]
        \centering
        \begin{tabular}{cccccc}
                \toprule
                \textbf{learning rate} & \textbf{epochs} & \textbf{warmup \%} & \textbf{scheduler} & \textbf{packing} & \textbf{target module} \\
                \midrule
                2e-5 & 1 & 0.03 & cosine & false & $W_q, W_v$ \\
                \bottomrule
        \end{tabular}
        \caption{Configuration for robustness experiments to different rank settings. Note that the batch size is set to 1 and the gradient accumulation steps is set to 100.}
        \label{tab:exp_robustness_rank_config}
\end{table*}


\end{document}